# Building a reordering system using tree-to-string hierarchical model

*Jacob DLOUGACH[1]   Irina GALINSKAYA[1]*
(1) Yandex School of Data Analysis, 16 Leo Tolstoy St., Moscow 119021, Russia
`jacob@yandex-team.ru, galinskaya@yandex-team.ru`

ABSTRACT

This paper describes our submission to the First Workshop on Reordering for Statistical Machine Translation. We have decided to build a reordering system based on tree-to-string model, using only publicly available tools to accomplish this task. With the provided training data we have built a translation model using Moses toolkit, and then we applied a chart decoder, implemented in Moses, to reorder the sentences. Even though our submission only covered English-Farsi language pair, we believe that the approach itself should work regardless of the choice of the languages, so we have also carried out the experiments for English-Italian and English-Urdu. For these language pairs we have noticed a significant improvement over the baseline in BLEU, Kendall-Tau and Hamming metrics. A detailed description is given, so that everyone can reproduce our results. Also, some possible directions for further improvements are discussed.

# 1   Introduction

As participants of the First Workshop on Reordering for SMT, we were required to build a system to reorder words in a source English sentence in such way, that it would match the order of words in a translation of that sentence into the target language (which could be Farsi, Urdu or Italian in our case).

After receiving the training data, we have noticed many common patterns in the sentences. For example, Farsi turned out to have constituent word order of "subject-object-verb" and noun order of (usually) "noun-modifier", which is different from English "subject-verb-object" and "modifier-noun" respectively. Considering a very small amount of training data (5000 sentences), we have decided that making a lexical-only model would be unreasonable, but such amount can still be enough for building a reliable syntax-based model (Quirk and Corston-Oliver, 2006), so we have decided to build such model with rules being automatically extracted from the training corpus.

# 2   Model training

## 2.1   Model description

We have used a model, often referred to as "tree-to-string" (Nguyen et al., 2008) to find the best reordering candidate. In this model some sequences of consecutive words (further referred to as word spans) are assigned syntax labels. These labels could either be syntax entities (like predicate) or part of speech tags. If the labels are induced by a syntax parse, they form a tree structure, i.e. if two labelled spans share common words, then one of them is enclosed inside the other one.

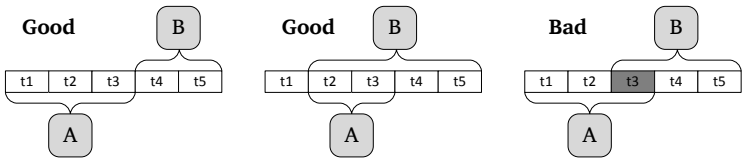

FIGURE 1 – EXAMPLES OF CORRECT AND INCORRECT SPAN LABELLING

For our purposes we can assume, that every word has its own label (i.e. part of speech tag). The model uses an assumption that we may assemble the reordering for the whole sentence from permutations of its syntax blocks (Hwa et al., 2002). More formally, we can describe this process using synchronous context-free grammar (further abbreviated as SCFG) (Chiang, 2007). Let's say that each of the possible labels is matched by a class within the grammar. Then the rules describing expansions of non-terminals will define a reordering *iff* there is a perfect matching between symbols on the source side of the rule and the target side (i.e. matching symbols should strictly coincide). It should be noted, that such expansion can include both terminals (single words) and non-terminals (syntax classes).

## 2.2 Dropping words

However, in the data that we have been provided with, some words from the source sentences may have been dropped and, thus, there were no words from the target sentences matching them. At this point, we have found two ways of adjusting the initial model to account for this peculiarity of the data. One possible approach is to remove some words after the reordering without changing the model itself. Another approach is to allow certain deletions inside the rules. Since we have had an "a priori" knowledge about the words that are to be dropped, we can enable the decoder to use this knowledge, so that the language model can estimate translation hypotheses become more precisely. We have assumed that only the syntax properties of dropped words matter for reordering of the rest of the sentence, so we have simply substituted these words with a special symbol that is guaranteed not to occur anywhere else in the data sets. The second approach has demonstrated significantly better performance on the development set, so we have decided not to include the results of the first one into our final submission.

## 2.3 Data preparation

The training set required some pre-processing for Moses to read it. We needed to convert the parse file into XML format accepted by Moses and also to provide the alignments between source and target sentences. The alignments have been derived directly from the data, whereas the XML was obtained from the parse escaping all special symbols and then applying the following substitution rules:

| `[`*{class}* | `<tree label="`*{class}*`">` |
|---|---|
| *{class}*`]` | `</tree>` |
| `{word}_{pos}` | `<tree label="`{pos}`">` {word} `</tree>` |

TABLE 1 – SUBSTITUTION RULES FOR PROCESSING THE PARSE

## 2.4 Training steps

Moses training pipeline consists of nine steps:

1. Prepare data
2. Run `GIZA++`
3. Align words
4. Get lexical translation table
5. Extract phrases
6. Score phrases
7. Build lexicalized reordering model
8. Build generation models
9. Create configuration file

In our case the data have been prepared separately and the alignments were given explicitly, so it was not required for us to run the first three steps. Also since we are training a hierarchical model, step 7 (lexicalized reordering model) is not applicable, and since we know exact translations of each word, it isn't reasonable to build a generation model either. Therefore we are only left with steps 4-6 and 9. It is worth noting, though, that step 4 can be done separately as well, because single words are always left unchanged during the translation process (except for special symbol standing for dropped words), and step 9 isn't really configurable, so further we will only focus on extraction and scoring of the rules.

### 2.4.1 Extraction

Extraction has been carried out using an *extract-rules* tool in Moses. Since default parameters in this tool are tuned assuming phrase-based translation, we have needed some adjustments. Here is the list of used non-default parameters:

| Parameter | Value | Comments |
|---|---|---|
| GlueGrammar | N/A | This parameter enables creation of rules to glue any two spans together without changing their order. When no rules can be applied, this one will always guarantee that at least one translation will be produced. |
| MinHoleSource | 1 | Default value is 2, which is good for hierarchical models, but too strict for syntax models. |
| MaxSymbolsSource | 4 | Greater values have proved to slow down the process of rule extraction and scoring too much. |
| MaxSpan | 999 | This means that we can extract rules spanned over the whole sentence. |
| MaxNonTerm | 4 | Default is 2, and we actually want to generate rules where all symbols could be non-terminals. |
| NonTermConsecSource | N/A | This allows two non-terminals on source side to appear adjacent to each other. |
| MinWords | 0 | This specifies the minimum number of terminals. Reasons for selection of this value are the same as in MaxNonTerm parameter. |

TABLE 2 – *EXTRACT* TOOL CONFIGURATION

Here are some examples of the extracted rules:

| Source phrase | Target phrase | Alignment |
|---|---|---|
| [ADJP][X] [NN][X] [NP] | [NN][X] [ADJP][X] [X] | 1-0 0-1 |
| having political [NNS][X] [VP] | having [NNS][X] political [X] | 0-0 2-1 1-2 |
| $ [CD][X] billion [QP] | [CD][X] billion $ [X] | 1-0 2-1 0-2 |

TABLE 3 – EXAMPLES OF EXTRACTED RULES

### 2.4.2 Scoring

Rules are then assigned weights using *score* tool. In order to be able to restore the alignment of the target phrase into the source phrase, we have specified "`WordAlignment`" flag (otherwise it would only print alignments of the non-terminals). Also we have decided to utilize Good-Turing frequency estimation (Good, 1953) due to low amount of available parallel sentences and the resulting data sparseness.

### 2.4.3 Language model

We have decided to build a simple 3-gram language model based on the target sentences as a corpus using IRSTLM toolkit (Federico, Bertoldi and Cettolo, 2008). The necessary steps exactly follow Moses tutorial on building a baseline system (Koehn, 2012). Briefly speaking, we have added sentence boundary symbols and have counted n-grams with Kneser-Ney smoothing (Chen and Goodman, 1996).

## 3 Decoding

Decoding has been performed using a chart decoder, implemented in Moses. Data preparation has involved building an XML representation of the parse tree, as in section 2.3.

### 3.1 Printing alignments

At the time we started carrying out the task, alignments output didn't work in the chart decoder: even though the corresponding option could be specified, the decoder would fail at loading time if word alignments were present in the rule table. It turned out that the decoder had relied heavily on the alignments being listed for non-terminals only, so the source code needed some enhancements to lift this restriction.

Then in order to print the alignments for a given translation we have recursively built alignments for each constituting hypothesis. Also we have needed to pay some attention to the unknown words, because the alignments would be explicitly set for them, which is always "0-0" assuming that words in the sentences are zero-indexed.

Since the option to print the alignments in chart decoder was highly demanded by the community, these changes have been integrated into the public Moses repository.

### 3.2 Decoding parameters

Since we have needed to generate the best possible translations, we have decided to lift most of the constraints in the decoder. Also we have manually added an entry into the rules table in order to delete the words that shouldn't be present in the target sentence (if we are substituting all these words with a special symbol as described in section 2.1). The parameters that we have changed from the default configuration, generated by training pipeline, are listed in Table 4.

| Parameter name | Value | Comment |
| --- | --- | --- |
| `ttable-limit` | 0 | Lifts the constraint on number of possible load translations per source phrase in rules table. |
| `cube-pruning-pop-limit` | 100000 | Number of top hypotheses to consider for each span. |
| `max-chart-span` | 1000, 1000 | Allows each rule to span across literally the whole sentence, thus enabling the decoder to move words from the beginning to the very end of even a long sentence. |

TABLE 4 – DECODER CONFIGURATION

## 4 Tuning

### 4.1 Technics

We have performed the tuning with the tools coming with Moses: MERT (Och, 2003) and MIRA (Venkatapathy and Joshi, 2007), both using only BLEU score for optimization. The tuned weights have corresponded to one feature in the language model and five features in the translation model (descriptions of each specific feature in the translation model can be found in Moses tutorial). It's worth noting, that the default value of using 100 best translations on each step hasn't been very efficient, because Moses has tended to generate 100 absolutely equal translations of one sentence using different rules and, for some reason, hasn't merged them while decoding, so we have used a limit of 2000. First, we have trained a model with removal of unneeded words after the translation process, and tuned it with MERT. However, when we decided to remove the words during decoding, all suggested metrics (which will be discussed further) have shown an increase of reordering quality on the development set even without any further tuning.

The tuning for the second model was not converging when MERT was used (actually, it seemed to be oscillating heavily), so we have utilized MIRA, which has recently been integrated into Moses. The changes occurring at every iteration have become less dramatic than with MERT, but on the other side number of iterations, required to get some stable result, has increased. The quality actually increased, but only by a small margin.

### 4.2 Analysis

Three of the features in translation model have been assigned negative weights. Since this is a rather strange event, we have tried to provide some explanation for it. One of those features corresponds to glue rules. Since glue rules actually have positive feature scores, it's pretty reasonable for them to be assigned negative weights, since their usage during translation results in unchanged order regardless of other rules. Another negative weight corresponds to phrase penalty. This means that decoder should attempt to use as few rules as possible, like in phrase-based translation, where using longer phrases would provide more reliable translation. The third weight is inverse phrase translation probability (conditional probability of source phrase provided the target phrase). While this could seem really strange for phrase-based translation, in syntax-based translation models having negative weight assigned to inverse translation probability results in taking additional syntax-based language model as a supplementary feature:

$$P_{final}(f|e) \propto p_{tm}(e|f)^{\lambda_1} \cdot p_{tm}(f|e)^{\lambda_2} \cdot p_{lm}(f)^{\lambda_3} = p_{tm}(e|f)^{\lambda_1+\lambda_2} \cdot \left(\frac{p_{tm}(f)}{p_{tm}(e)}\right)^{\lambda_2} \cdot p_{lm}(f)^{\lambda_3}$$

In this formula $p_{tm}$ stands for probabilities estimated by translation model, while $p_{lm}$ stands for language model approximation. $f$ is the sentence with Farsi word order, and $e$ is the source English sentence. Note that this differs from the notation commonly used in other statistical machine translation works, where $f$ would be source language and $e$ would be target language.

In our case $\lambda_1$ is negative, but $\lambda_2$ and $\lambda_3$ are positive. Moreover, $\lambda_1 + \lambda_2$ is positive too. Notice that $p_{tm}(f)$ is can be treated as another language model, which is syntax-aware. Therefore, we can come to a conclusion, that inverse translation probability actually would be assigned a positive weight if our language model was syntax-based. Also, cumulative weight of the language model is equal to $\lambda_2 + \lambda_3$, which in our case is approximately 1.5 times higher than cumulative weight of the translation model – $\lambda_1 + \lambda_2$. Thus, we can conjecture that having a better language model could considerably increase the quality of our reordering.

## 5 Evaluation

Model training has been carried out on a 3 sets of 5000 English sentences each (all sets corresponding to different language pairs). Regardless of the target languages we followed exactly the same procedure for model training as described above in section 2. Both development and testing sets have consisted of 500 sentences each. As a baseline we have taken the unaltered word order.

### 5.1 Metrics

Three metrics have been used for the evaluation: BLEU, Kendall's tau distance, and Hamming distance (Birch and Osborne, 2010). It should be noted though, that the last two are measured in fraction remaining to maximum value (i.e. if distance is 0 the metric would be 1.0, and if distance is maximal possible the metric equals 0.0).

### 5.2 Results for development set

| Model | English – Farsi | English – Italian | English – Urdu |
|---|---|---|---|
| | BLEU (%) / Kendall tau / Hamming | | |
| Baseline | 51.29 / 0.761 / 0.435 | 69.0 / 0.867 / 0.723 | 39.5 / 0.52 / 0.274 |
| Delete words after translation; tuned with MERT | 67.1 / 0.795 / 0.532 | N/A | N/A |
| Delete words during translation; tuned with MERT | 69.5 / 0.805 / **0.567** | N/A | N/A |

| Model | English – Farsi | English – Italian | English – Urdu |
|---|---|---|---|
| Delete words during translation; tuned with MIRA | **69.8 / 0.807 /0.567** | **78.3 / 0.884 / 0.779** | **55.7 / 0.649 / 0.431** |

TABLE 5 – SCORES FOR THE DEVELOPMENT SET

As you can see, although we have only optimized BLEU, all other metrics increase at the same time. The results for different models are only included for English-Farsi because it has been our primary language pair for this shared task, while English-Italian and English-Urdu results have only qualified as post-submission experiments.

### 5.3 Results for testing set

For the testing set only BLEU scores are known.

| Model | English – Farsi | English – Italian | English – Urdu |
|---|---|---|---|
| Baseline | 50.0 | 66.4 | 39.0 |
| Delete words during translation; tuned with MERT | 65.24 | N/A | N/A |
| Delete words during translation; tuned with MIRA | **65.56** | **76.65** | **55.79** |

TABLE 6 – SCORES FOR THE TESTING SET

**Conclusion and perspectives**

We have managed to build a reordering system without any prior knowledge of the target language. The model has been built with Moses training pipeline and then has been applied to the testing data using chart decoder. We could observe a significant increase in all quality metrics comparing to a simple baseline (not reordered sentences).

Under the time constraints of the workshop, we haven't been able to try all of the options, so there are some ways for improvements. First of all it may be worth changing some of the parameters in learning, such as length of generated rules or smoothing options. Another way is to relax syntax constraints to allow more aggressive reordering when the parse tree is very sparse (i.e. some nodes have many children). As far as we could see after manually inspecting errors in our reordering, this will potentially boost the quality of reordering, however it will require some changes in Moses training scripts. Also, our analysis shows, that it may be very useful to utilize a better language model during decoding.

# References

Birch, A. and Osborne, M. (2010). LRscore for Evaluating Lexical and Reordering Quality in MT. *Proceedings of the Joint Fifth Workshop on Statistical Machine Translation and MetricsMATR*, Uppsala, Sweden, 327-332.

Chen, S.F. and Goodman, J. (1996). An empirical study of smoothing techniques for language modeling. *Proceedings of the 34th annual meeting on Association for Computational Linguistics*, Stroudsburg, PA, USA, 310-318.

Chiang, D. (2007). Hierarchical Phrase-Based Translation, *Computational Linguistics*, vol. 33, no. 2, June, pp. 201-228, Available: ISSN: 0891-2017 DOI: 10.1162/coli.2007.33.2.201.

Chiang, D., Lopez, A., Madnani, N., Monz, C., Resnik, P. and Subotin, M. (2005). The Hiero Machine Translation System: Extensions, Evaluation, and Analysis. *Proceedings of Human Language Technology Conference and Conference on Empirical Methods in Natural Language Processing*, Vancouver, British Columbia, Canada, 779-786.

Federico, M., Bertoldi, N. and Cettolo, M. (2008). IRSTLM: an open source toolkit for handling large scale language models. *INTERSPEECH*, 1618-1621.

Fox, H. (2002). Phrasal Cohesion and Statistical Machine Translation. *Proceedings of the Conference on Empirical Methods in Natural Language Processing (EMNLP)*, Philadelphia, 304-311.

Good, I.J. (1953). The population frequencies of species and the estimation of population parameters, *Biometrika*, vol. 40(3 and 4), pp. 237-264.

Hoang, H. and Koehn, P. (2008). Design of the moses decoder for statistical machine translation. *Software Engineering, Testing, and Quality Assurance for Natural Language Processing*, Stroudsburg, PA, USA, 58-65.

Hoang, H. and Koehn, P. (2010). Improved translation with source syntax labels. *Proceedings of the Joint Fifth Workshop on Statistical Machine Translation and MetricsMATR*, Stroudsburg, PA, USA, 409-417.

Hoang, H. and Lopez, A. (2009). A unified framework for phrase-based, hierarchical, and syntax-based statistical machine translation. *In Proceedings of the International Workshop on Spoken Language Translation (IWSLT*, 152-159.

Huang, L., Knight, K. and Joshi, A. (2006). Statistical Syntax-Directed Translation with Extended Domain of Locality. *5th Conference of the Association for Machine Translation in the Americas (AMTA)*, Boston, Massachusetts.

Hwa, R., Resnik, P., Weinberg, A. and Kolak, O. (2002). Evaluating Translational Correspondence using Annotation Projection. *Proceedings of the 40th Annual Meeting of the Association of Computational Linguistics (ACL)*.

Koehn, P. (2012). Moses: Statistical Machine Translation System. User Manual and Code Guide. http://www.statmt.org/moses/manual/manual.pdf. Accessed on 1 October 2012.

Koehn, P. and Hoang, H. (2007). Factored Translation Models. *Proceedings of the 2007 Joint Conference on Empirical Methods in Natural Language Processing and Computational Natural Language Learning (EMNLP-CoNLL)*, 868-876.

Koehn, P., Hoang, H., Birch, A., Callison-Burch, C., Federico, M., Bertoldi, N., Cowan, B., Shen, W., Moran, C., Zens, R., Dyer, C.J., Bojar, O., Constantin, A. and Herbst, E. (2007). Moses: Open Source Toolkit for Statistical Machine Translation. *Proceedings of the 45th Annual Meeting of the Association for Computational Linguistics Companion Volume Proceedings of the Demo and Poster Sessions*, Prague, Czech Republic, 177-180.

Li, C.-H., Li, M., Zhang, D., Li, M., Zhou, M. and Guan, Y. (2007). A Probabilistic Approach to Syntax-based Reordering for Statistical Machine Translation. *Proceedings of the 45th Annual Meeting of the Association of Computational Linguistics*, Prague, Czech Republic, 720-727.

Li, C.-H., Zhang, H., Zhang, D., Li, M. and Zhou, M. (2008). An Empirical Study in Source Word Deletion for Phrase-Based Statistical Machine Translation. *Proceedings of the Third Workshop on Statistical Machine Translation*, Columbus, Ohio, 1-8.

Nguyen, T.P., Shimazu, A., Ho, T.B., Nguyen, M.L. and Nguyen, V.V. (2008). A Tree-to-String Phrase-based Model for Statistical Machine Translation. *CoNLL 2008: Proceedings of the Twelfth Conference on Computational Natural Language Learning*, Manchester, England, 143-150.

Och, F.J. (2003). Minimum Error Rate Training in Statistical Machine Translation. *Proceedings of the 41st Annual Meeting of the Association for Computational Linguistics*, 160-167.

Och, F.J. and Ney, H. (2003). A Systematic Comparison of Various Statistical Alignment Models, *Computational Linguistics*, vol. 29, no. 1, pp. 19-52.

Papineni, K., Roukos, S., Ward, T. and Zhu, W.-J. (2001). BLEU: a Method for Automatic Evaluation of Machine Translation. Tech. rep. *IBM Research Report*.

Quirk, C. and Corston-Oliver, S. (2006). The impact of parse quality on syntactically-informed statistical machine translation. *Proceedings of the 2006 Conference on Empirical Methods in Natural Language Processing*, Sydney, Australia, 62-69.

Venkatapathy, S. and Joshi, A.K. (2007). Discriminative word alignment by learning the alignment structure and syntactic divergence between a language pair. *Proceedings of the NAACL-HLT 2007/AMTA Workshop on Syntax and Structure in Statistical Translation*, Stroudsburg, PA, USA, 49-56.

Yamada, K. and Knight, K. (2001). A Syntax-Based Statistical Translation Model. *Proceedings of the 39th Annual Meeting of the Association of Computational Linguistics (ACL)*.